\documentclass[11pt]{article}
\usepackage{geometry} 
\usepackage{algorithm2e}

\usepackage{amsmath,mathrsfs,amssymb,amsthm,amscd}
\usepackage{color,epsfig,latexsym,graphicx,eucal,layout,fancyhdr}
\usepackage[]{fontenc}
\usepackage[all]{xy}
\usepackage{hyperref}
\usepackage{graphics}
\usepackage{a4wide}
\usepackage{palatino}
\usepackage{setspace}
\usepackage{parskip}
\newcounter{EQNR}
\newcommand{\NN}{\field{N}}
\newcommand{\field}[1]{\ensuremath{\mathbb{#1}}}

\let\oldtext\text
\renewcommand{\text}[1]{{\textbf{\small \oldtext{#1}}}}

\title{Automated timetabling for small colleges and high schools using huge integer programs}
\author{Joshua S. Friedman\footnote{The views expressed in this article are the author's own and not those of the U.S. Merchant Marine Academy,
the Maritime Administration, the Department of Transportation, or the United States government.}}

\date{\today} 

\begin{document}
\maketitle

\abstract{We formulate an integer program to solve a highly constrained academic timetabling problem at the United States Merchant Marine Academy. The IP instance that results from our real case study  has approximately both 170,000 rows and columns and solves to optimality in 4--24 hours using a commercial solver on a portable computer (near optimal feasible solutions were often found in 4--12 hours). Our model is applicable to both high schools and small colleges who wish to deviate from group scheduling. We also solve a necessary preprocessing student subgrouping problem, which breaks up big groups of students into small groups so they can optimally fit into small capacity classes.}

\tableofcontents

\section{Introduction}
Solving real world academic timetabling problems can quite difficult. Many heuristic  approaches have been developed, including graph coloring methods using Kempe exchanges \cite{muh, muhwanka}, Max-SAT solvers \cite{acha}, genetic algorithms \cite{abramson,domros}, ant colony optimization \cite{socha} , tabu search \cite{lu,souza}, and hybrid approaches \cite{jonasson}. See also the surveys \cite{schaerf,rohini,babaei}.

In theory, mixed integer programs can give exact solutions, however in practice they can be so large that they are computationally unsolvable. Never-the-less, for smaller and medium sized instances, integer programming has been used successfully \cite{pillay,lach,daskalaki,helber,birbas,burke,kristiansen}.

We solve the academic timetabling problem at the United States Merchant Marine Academy (USMMA).
The problem we consider seems to be a more complex, real world, timetabling problem than others that have been solved by integer programming techniques. For most colleges, a curriculum timetabling model is appropriate, where one makes sure that required courses within a curriculum do not clash, while students are free to fill in their schedule with a wide choice of electives. Our situation is much less flexible. Though we are a fully accredited higher education academic institution, our scheduling requirements are more like a high school's, albeit one that offers six majors, has 900 students, roughly 95 percent of all classes are required, and students must graduate in 4 years with approximately 160 credits. In addition, both our students and professors can have very dense schedules. Finally, we must guarantee that all students can enroll in all of their required courses. Our approach is student based, that is, the primary goal is for all students to be able to enroll in all the courses they need.

To very quickly give a sense of our problem, we give a simple example of the type of input our algorithm requires. The following data is overly simplified and not valid. 

The first set of data is the courses requested by students. Each row represents a group of students. For example, in row 1, there are 37 students and all of them needs to take the four courses: CALC1, STAT2, COMP1, and PHYS1. Though they all need the same courses, they will not necessarily be in the same sections.\footnote{Sections are the events that are scheduled, and there can be multiple sections of each course.} 

\begin{tiny}
\begin{tabular}{ | l | l | l | l | l | l | l | l | }
\hline
	Student Groups &  &  &  &  &     \\ \hline
	 &  &  &  &  &     \\ \hline
	Group Number & Size of Group & 1st course needed & 2nd course needed & 3rd course needed & 4th course needed   \\ \hline
	1 & 37 & CALC1 & STAT2 & COMP1 & PHYS1   \\ \hline
	2 & 42 & ENGL1 & PHYS1 & PHYS1LAB & STAT2     \\ \hline
	3 & 14 & CALC1 & PHYS1 & PHYS1LAB & ENGL1     \\ \hline
	4 & 19 & COMP1 & PHYS1 & PHYS1LAB & STAT2     \\ \hline
	5 & 39 & CALC1 & PHYS1 & PHYS1LAB & GEOM1     \\ \hline
	6 & 1 & ENGL1 & CHEM1 & CHEM1LAB & COMP1    \\ \hline
\end{tabular}
\end{tiny}

\bigskip

Second, we have sections (these are the actual classes we schedule), that is multiple copies of the same course offered. Note that the capacity column can not be exceeded, so the groups above will need to be partitioned into subgroups (see \S\ref{Secsubgroup}) in such a way so all students fit into the classes that they are required to take.    

\bigskip

\begin{tiny}
\begin{tabular}{ | l | l | l | l | l | l | l | l | }
\hline
	Sections Offered &  &  &  &  &    \\ \hline
	 &  &  &  &  &    \\ \hline
	Prof. Last Name & Course & Periods & Lab & Capacity & Room type    \\ \hline
	Gauss & CALC1 & 3 & N & 26 & CLASSROOM   \\ \hline
	Gauss & GEOM1 & 3 & N & 26 & CLASSROOM   \\ \hline
	Gauss & CALC1 & 3 & N & 26 & CLASSROOM  \\ \hline
	Riemann & CALC1 & 3 & N & 26 & CLASSROOM   \\ \hline
	Riemann & STAT2 & 3 & N & 17 & CLASSROOM   \\ \hline
	Turing & COMP1 & 4 & N & 22 & CLASSROOM   \\ \hline
	Turing & COMP1 & 4 & N & 22 & CLASSROOM   \\ \hline
	Einstein & PHYS1 & 3 & N & 21 & CLASSROOM   \\ \hline
	Einstein & PHYS1LAB & 2 & Y & 21 & PHYSLAB   \\ \hline
	Bohr & PHYS1 & 3 & N & 21 & CLASSROOM   \\ \hline
	Pauli & PHYS1LAB & 2 & Y & 21 & PHYSLAB   \\ \hline
	Curie & CHEM1 & 3 & N & 25 & CLASSROOM   \\ \hline
	Curie & CHEM1LAB & 2 & Y & 15 & CHEMLAB   \\ \hline
	Austen & ENGL1 & 3 & N & 20 & LECTUREHALL  \\ \hline
	Austen & ENGL1 & 3 & N & 20 & LECTUREHALL  \\ \hline

\end{tabular}
\end{tiny}

\bigskip 
Third, we have a simple database of rooms. Note that the room type column corresponds to the sections, above.
\bigskip

\begin{tiny}
\begin{tabular}{ | l | l | l | }
\hline
Rooms  Available & &  \\ \hline
	Room & cap & Room type   \\ \hline
	F101 & 30 & CLASSROOM   \\ \hline
	F102 & 30 & CLASSROOM   \\ \hline
	F103 & 30 & CLASSROOM   \\ \hline
	F310 & 21 & PHYSLAB   \\ \hline
	F339 & 15 & CHEMLAB   \\ \hline
	B101 & 30 & LECTUREHALL   \\ \hline
	B102 & 30 & LECTUREHALL   \\ \hline
\end{tabular}
\end{tiny}

\bigskip
Finally, we have the professor time requests. Note that a zero means the professor prefers not to teach at that time (7 periods per day, 5 days per week).

\begin{tiny}
\begin{tabular}{ | l | l | l | l | l | l | l | l | }
\hline
	Professor Time Preferences &  &  &  &  &  &  &  \\ \hline
	 &  &  &  &  &  &  &  \\ \hline
	Prof Einstein &  &  &  &  &  &  &  \\ \hline
	 & 1 & 2 & 3 & 4 & 5 & 6 & 7 \\ \hline
	M & 0 & 0 & 0 & 0 & 0 & 0 & 0 \\ \hline
	T &  &  &  &  &  &  &  \\ \hline
	W &  &  &  &  &  &  &  \\ \hline
	R &  &  &  &  &  &  &  \\ \hline
	F & 0 & 0 & 0 & 0 & 0 & 0 & 0 \\ \hline
	 &  &  &  &  &  &  &  \\ \hline
	Prof Gauss &  &  &  &  &  &  &  \\ \hline
	 & 1 & 2 & 3 & 4 & 5 & 6 & 7 \\ \hline
	M & 0 &  &  &  &  &  & 0 \\ \hline
	T & 0 &  &  &  &  &  & 0 \\ \hline
	W & 0 &  &  &  &  &  & 0 \\ \hline
	R & 0 &  &  &  &  &  & 0 \\ \hline
	F & 0 &  &  &  &  &  & 0 \\ \hline
	 &  &  &  &  &  &  &  \\ \hline
	Prof Riemann &  &  &  &  &  &  &  \\ \hline
	 & 1 & 2 & 3 & 4 & 5 & 6 & 7 \\ \hline
	M & 0 & 0 &  &  &  & 0 & 0 \\ \hline
	T & 0 & 0 &  &  &  & 0 & 0 \\ \hline
	W & 0 & 0 &  &  &  & 0 & 0 \\ \hline
	R & 0 & 0 &  &  &  & 0 & 0 \\ \hline
	F & 0 & 0 &  &  &  & 0 & 0 \\ \hline
\end{tabular}
\end{tiny}
\bigskip

The problem we solve is to schedule all sections without any conflicts resulting, so that all students are  enrolled in all of their courses, and that we violate as few professor time requests as possible (as well as many other hard and soft constraints).

\subsection{Subgroup problem}
In addition to solving the standard assignment problems (assigning classes to rooms, classes to times) we also assign various groups of students to classes, so that they do not exceed the class's capacity. For example, suppose we have three groups of students of size $34,41,15$ all who need MATH101, PHYS101, and ENGL101 which have capacity 30,25,15 students, respectively.  We will need to offer three sections of MATH101 to handle the 90 students, four sections of PHYS101, and 6 sections of ENGL101. The problem that arrises is a hybrid assignment knapsack/bin packing problem. The first problem we must solve, a preprocessing problem, is to break up large groups into smaller groups in such a way that they can packed into the minimal number of courses needed. Though we can break up each group into singleton subgroups, the resulting IP would be computationally unsolvable as it has too many rows and columns.

\subsection{Overview of our algorithm} 
We first assume that enough sections are offered to meet demand. In \S\ref{Secsubgroup} we solve our subgroup problem. The idea is to use an integer program to assign all groups of students to sections of courses that they need. In this initial stage, we will not succeed because the groups are too large. The IP minimizes the overcapacity of the assignment and makes sure that it is possible for groups of students to be enrolled in courses with labs. Then a greedy approach looks for the most over-scheduled section and identifies the biggest group in that section. That group is split into two smaller groups so that one of the new groups will contain only the unscheduled students. Next we re-run the IP with the new, finer, groups, and repeat the process until all students fit into classes. It takes less than 50 iterations to terminate (under 1 minute to complete all iterations).  

Once we have subgroups that fit, we run our timetabling integer program (TIP). We \emph{do not} store the assignment of groups to students from the subgroup problem, even though it would speed up the TIP because it would be suboptimal: a priori, we don't know which groups should be combined to fit into a particular section without considering the global timetabling problem with all of its constraints. Incorporating the subgroup problem into the TIP, would probably not be computationally solvable. 

Our TIP schedules sections so that professors and rooms do not have conflicts. The groups are assigned sections in such a way that they are enrolled in the courses they need, and so that there are no group time conflicts. Since we actually solve a real problem, there are many constraints we need to deal with: groups who take classes with labs should have the same professor for both, a class and its lab shouldn't be back to back, labs should be scheduled contiguously and in the same room; and many soft constraints: no professor should be assigned to period 1 and period 7 on the same day, each professor should have at least one day off, and most professors should have classes on meeting days.

Our instance\footnote{Term 2 of the 2016-2017 academic year.} has the following key components: 652 students, 334 distinct sections, 126 professors, and 91 classrooms (many of which are specialty labs). 
The resulting IP is mostly binary, and it is huge. Before pre-solving, it has 178,496 rows, 175,594 columns, and 1,650,967 nonzero entries.  Both Gurobi 6.5 and 7.0 were able to solve the instance.\footnote{Gurobi 7.0 reduced solving times by at least 30 percent over Gurobi 6.5} Gurobi 7.0 solved our instance in  4--24 hours\footnote{The variation seems to be due to the random seed parameter that the solver uses. We solved our instance with various seeds.}, after tuning.  We also attempted to solve our instance with GLPK \cite{glpk} , SCIP \cite{scip}, CBC \cite{cbc}, and CPLEX \cite{cplex}, but were unsuccessful.  We used the GNU MathProg modeling language that comes with \cite{glpk} and solved the resulting LP-file with Gurobi. Our hardware was extremely modest: a Macbook Pro laptop computer, with a Core i7 (4 cores, 2 threads/core) 2.5 GHZ CPU and 16 GB RAM. We specified the solver to use seven execution threads. 

We also constructed a small artificial instance with 23,294 rows, 28,926 columns and 268,350 non-zeros. It usually solved in 5--20 minutes with Gurobi. However, when we assigned some groups a dense schedule with all 35/35 spots utilized, it took the solver many times longer to find the optimal solution.

We tested our model for the second trimester of the 2016-2017 academic year on real life data and verified correctness with the Registrar. In addition, we have recently tested our model on the third trimester and obtained optimal results in approximately 6--12 hours.

Great advances in mixed integer programming solver technology have occurred relatively\footnote{
"The field of mixed integer programming has witnessed remarkable improvements in recent years in the capabilities of MIP algorithms. Four of the biggest contributors have been presolve, cutting planes, heuristics, and parallelism" \mbox{(\url{gurobi.com/resources/getting-started/mip-basics}).}} recently, and allowed us to solve the problem in this way. In fact as mentioned roughly 7 years ago in \cite{lach} 

\begin{quote}...we report for the first time on solving the four original Udine instances to proven optimality but certainly not only due to the fact that they became rather easy for modern integer programming solvers.
\end{quote}

\subsection{Applications of our method}
High schools and small colleges do not have the resources that a large college may have. A small school may only offer two sections of calculus while students from three different majors may need the course. At high schools, students are placed in tracks where only certain combinations of courses may be possible, and thus there is little flexibility for individualized schedules. Small schools are stuck in a group scheduling mode. Our method could allow any high school (or middle school) to have much more flexible student schedules, and allow teachers to have greater control over their schedules.  More efficient timetabling could reduce the total number of sections needing to be offered. The big colleges are probably automated, to a certain degree. It is the small schools who are timetabling by hand, who have the most to gain.

\subsection{Timetabling at USMMA} 
The United States Merchant Marine Academy is a service academy with approximately 900 midshipmen (students). Scheduling of students, professors, and classrooms is currently being done by hand, and is both quite challenging and time consuming. The midshipmen are all full-time and have a highly constrained schedule. They must be enrolled in all of their required courses, and virtually every course they take is required. There are 35 periods each week (each period is 55 minutes) and it is quite common for large numbers of midshipmen to have 30-33 periods filled with required courses.

\begin{table}
\centering
\begin{tiny}

\begin{tabular}{|l|l|l|l|l|l|l|l|}
\hline
\begin{tabular}[c]{@{}l@{}}Group 38\\ Size = 12\end{tabular} & 1 & 2 & 3 & 4 & 5 & 6 & 7 \\ \hline
Monday & NAUT130 & BUSN110 & NASC100 & PEA125L & \begin{tabular}[c]{@{}l@{}}COMMON\\ HOUR\end{tabular} & NAUT110 & PHYS110 \\ \hline
Tuesday & PEA120L & HIST100 & NAUT120L & NAUT120L & BUSN110 & NAUT120 & PHYS110 \\ \hline
Wednesday &  & BUSN101 & PEA125L & NASC100 & PHSY110L & PHYS110L & HIST100 \\ \hline
Thursday & BUSN110 &  &  & MLOG120 &  & NAUT120 & PHYS110 \\ \hline
Friday & PEA120L & HIST100 &  &  & NAUT110 & NAUT120L & NAUT120L \\ \hline
\end{tabular}
\caption{ \small{Typical schedule of a freshman (plebe) at USMMA.} }
\end{tiny}
\end{table}

Currently students are split into groups, and timetabling is done group-wise (the only way to construct these timetables by hand). However, major problems occur when a student fails a course (or places out of a course) and leaves the common group schedule. 

Midshipmen attend the Academy for 11 months per year, for 4 years, with three trimesters per year. Each trimester is equivalent to a semester at other colleges. Midshipmen spend 3 trimesters on commercial ships, and in the remaining 9 trimesters they must take all the required courses for a B.S. degree, a U.S. Coast Guard license, and a commission in the U.S. Navy. They end up with roughly 160 credit hours. 

The midshipmen are partitioned into 48 groups: At any given time half of the sophomores and juniors are out to sea, depending on whether they are A-split or B-split. There are 2 splits, 6 academic majors, and 4 class years. Each semester there are approximately 36 regular groups of midshipmen who are on campus, hence 36 curriculums.  In addition there are exceptional groups of midshipmen who fall outside of the common group schedule. Though all members of a particular group need the same courses, they may not have identical timetables (if the group is large, it would be split up).

\subsection{Acknowledgment}
I would like to thank Lisa Jerry, Registrar U.S. Merchant Marine Academy, for both providing two trimesters worth of real academic scheduling data, and for spending many hours putting the data into a useful form that my algorithms could process. I could not have done the project without her support. I would also like to thank Maribeth Widelo for assisting with the real data and for providing valuable feedback from her expertise in academic scheduling. Thanks are also due to Dr. Mark Hogan for supporting the project.

\section{Definitions}

\subsection{Rooms}
Let $R$ denote the set of rooms, and let $\text{RT}$ denote the set of \emph{room types.} Each room $r$ is assigned a room type, $\text{type}(r) \in \text{RT}.$ Room types, can be, for example, general classrooms in a certain building, laboratories, specialty classrooms,\dots 

Each room has a capacity $\text{cap}(r).$

\subsection{Time and Day}
Each day has seven 55 minute periods reserved for classes $T = \{1,2,3,4,5,6,7\}$ (lunch falls between period 4 and 5) and classes are scheduled five days per week on the days $D = \{M,T,W,R,F\}.$

\subsection{Professors}
Let $P$ denote the set of professors. To each professor $p \in P$ we associate a $5\times7$ matrix of availability $\text{avail}(p),$ where $\text{avail}(p)_{dt} = 1$ if professor $p$ can teach on day $d$ and time $t;$ and if professor $p$ can not teach, then $\text{avail}(p)_{dt} = 0,-1,-2$ where $0$ denotes they prefer not to teach, $-1$ it is important that they do not teach, and $-2$ they absolutely can not teach, at that time and day. 

If professor $p$ is an adjunct, then $\text{adj}(p) = \text{True},$ otherwise $\text{False}.$ At USMMA, adjuncts get the highest priority for time availability.

\subsection{Courses}
Let $C$ denote the set of courses offered
$$C = \{ \text{MATH101, MATH120, MATH210, BUSN101, PHYS110,} \dots, \}.$$  
Note that multiple copies of each course are generally offered. Each course $c$ requires a number of periods which we denote by $\text{periods}(c) \in \{1,2,3,4,5\}.$ For us, a course is just a name with a number of periods. One could also define a course as a set of sections (\S\ref{subSec}). It is the sections that are actually scheduled.

\subsection{Sections} \label{subSec}
Let $S$ denote the set of sections being offered. Sections are the objects which are actually scheduled.  Each element $s \in S$ has the following parameters (maps): 
\begin{enumerate}
\item $\text{prof}(s) \in P,$ the professor teaching section $s.$

\item $\pi(s) \in C,$ the course associated with section $s.$ For example, if $s$ is a section of MATH101, $\pi(s) = \text{MATH101}.$ 

\item $\text{periods}(s) \in \{1,2,3,4,5\},$ the number of time slots (number of periods) needed to be scheduled per week. Note that we overload the $\text{periods}(\cdot)$ function so that $\text{periods}(\pi(s)) = \text{periods}(s).$

\item $\text{lab}(s) \in \{\text{True,\,False}\}.$ If section $s$ requires all of its scheduled periods to be scheduled consecutively,  in the same room, and without lunch interrupting the lab, then $\text{lab}(s) = \text{True},$ otherwise $\text{False}.$   

\item $\text{cap}(s) \in \NN = \{1,\dots\} \cup \{ 0 \}$ the capacity of $s,$ how many students $\text{prof}(s)$ allows in his/her section. If we want to reserve a room for a professor, but not enroll students, we set the capacity to zero.

\item $\text{final}(s) \in F = \{\emptyset\} \cup C,$ the final exam associated to section $s.$

\item $\text{roomtype}(s) \in \text{RT},$ the subset of rooms that are appropriate for section $s.$ That is, room $r$ is appropriate if $\text{type}(r) = \text{roomtype}(s).$

\item $\text{labtie}(s) \in \{\emptyset\} \cup \NN.$ Suppose $s_1,s_2 \in S$ with $\text{lab}(s_2) = \text{True},$ and $\text{lab}(s_1) = \text{False}.$  Then all students enrolled in section $s_2$ must also be enrolled in $s_1$ if and only if $\text{labtie}(s_1) =\text{labtie}(s_2).$ 

\item For $i=1\dots 5,$ $\text{mandate}_i(s) \in (D \times T ) \cup (\emptyset,\emptyset), $  requring that $s$ is scheduled for a specific day and time. For example, if $\text{mandate}(s) = ((T,5), (W,3), (F,5), (\emptyset,\emptyset), (\emptyset,\emptyset))$ then we are insisting that $s$ is scheduled Tuesday 5th period, Wednesday 3rd, and Friday 5th.
\item For $i \in \{1,\dots 6\},$ $\text{coprof}_i(s) \in P \cup \{\emptyset\},$ for possible alternate professors who must be available to teach section $s.$

\item $\text{adjunct}(s) \in \{\text{True,\,False}\},$ if a section is taught by an adjunct professor, they get the highest priority because they often have very limited availability.

\item $\text{link}(s) \in \{\emptyset \} \cup \NN.$ Two sections $s_1,s_2$ must be taught at identical days and times if and only if  $\text{link}(s_1)=\text{link}(s_2).$
\end{enumerate}

\subsection{Groups} \label{secGroups}
Let $G$ denote the set of groups of students. A group $g \in G$ is a set of students who all need the same exact courses, though not necessarily the same sections. Typically each $g$ represents students who are in the same major and in the same class year. \emph{If a student has an exceptional curriculum, they may be in a group of size one.} Groups can be small to allow for great flexibility. However, the more groups that are present, the longer it will take for an optimal solution to be found.

Each $g \in G$ has a curriculum, that is a set of courses they must take, which we denote by $C_g \subset C.$  For each $c \in C_g$ we must find a unique section $s \in S$ with $\pi(s)=c$ to enroll the entire group $g.$ 

The size of $g$ is denoted by $\text{size}(g) \in \NN.$

As mentioned previously, the \emph{Subgroup Problem} needs to be solved (see \S\ref{Secsubgroup}) which will result in slightly smaller groups. This problem is related to the Student Sectioning Problem \cite{rud} and \cite[p. 16]{muh}.

\section{The Constraints and Objective}
The constraints are separated into hard and soft constraints. 
\subsection{Hard Constraints}
\begin{enumerate}
\item Each section $s$ must be scheduled for exactly $\text{periods}(s)$ time slots per week.

\item If two sections $s_1,s_2$ have $\text{link}(s_1)=\text{link}(s_2),$ then they must be taught at identical days and times.

\item For each $d \in D, t \in T, r \in R$ at most one section $s$ can be assigned (one class at a time in each room).

\item Each section that has mandated times must be scheduled accordingly.

\item Each section that is not a lab can meet at most one time per day.

\item Each section that is a lab must have all its meetings on the same day, in the same room, contiguously scheduled, and without lunch falling within the allotted times.

\item For each $p \in P$ the set of sections $s$ with $\text{prof}(s)=p$ must be scheduled at disjoint times (no collisions allowed), and they must all be scheduled.  

\item For each $s \in S, p \in P$ with $\text{coprof}_i(s)=p$ for some $i,$ each section $s_k$ with $\text{prof}(s_k)=p$ must be scheduled so professor $p$ can attend section $s.$ In other words, if $p$ has regular sections, they can not make it impossible for him/her to attend sections where he/she co-teaches. 

\item For each group $g\in G,$ and each $c \in C_g,$ group $g$ must be scheduled for exactly one section $s \in \pi^{-1}(c).$ Note that we are not preassigning groups to sections. 

\item For each $g \in G,$ the sections that $g$ is enrolled in must be scheduled at disjoint times (no collisions).

\item If $s_0 \in S$ is a non-lab, and $s_1 \in S$ is a lab, and $\text{labtie}(s_0) = \text{labtie}(s_1) \neq \emptyset,$ then all groups enrolled in $s_1$ must also be enrolled in $s_0.$ This constraints keeps labs and lectures together, with the same professor, if requested. 

\item For each section $s,$ the number of students enrolled should not exceed $\text{cap}(s).$ We sometimes make this a soft constraint, but impose a big penalty for over capacity. We make sure that the rooms requested for $s$ have enough capacity, so we do not explicitly check room capacity. 

\item If a group $g$ is scheduled for two sections $s_1, s_2$ with $\text{labtie}(s_0) = \text{labtie}(s_1),$ then for each day, at most four hours should be scheduled, combined, for both $s_1$ and $s_2.$ In other words, we don't want a four hour lab scheduled on the same day as the lecture it is tied to.

\item  If a group $g$ is scheduled for two sections $s_1, s_2$ with $\text{labtie}(s_0) = \text{labtie}(s_1),$ then there should be at least a one period break (a lunch break is acceptable) between the two sections. In other words, a lab should not start right before or after lecture if they are tied together. 

\end{enumerate}

\subsection{Soft Constraints}

\begin{enumerate}
\item Each section $s$ that is not a lab, with $\text{periods}(s)=3,$ should not be scheduled in three consecutive days.

\item Each section $s$ that is not a lab, with $\text{periods}(s)=2$ should not be scheduled in two consecutive days.

\item Each professor must have at least one day without teaching.

\item On any given day, a professor should not teach both first period and seventh period.

\item All full-time professors (those teaching 9 or more hours per week) should be assigned teaching duties on Tuesdays (for weekly department meetings).

\item For each $p \in P,$ if $\text{avail}(p)_{dt} \leq 0,$ then do not schedule any section with $\text{prof}(s) = p,$ or $\text{coprof}_i(s) = p,$ at time and day $(d,t).$

\end{enumerate}

\subsection{The Objective}
The objective is to minimize the number of soft constraints violated. Each soft constraint has a weight. Violating the time preferences of an adjunct professor has the most penalty, and scheduling a three period class over only three days has the least. More details are in a later section.

\section{The Subgroup Problem} \label{Secsubgroup}
Recall \S\ref{secGroups} where we defined the set of groups $G,$ where each $g \in G$ is a set of students, all who need the same set of courses $C_g.$ In this section we solve the Subgroup Problem.

Starting with $G$ we define a refinement $G^{(1)}$ of $G.$ 

For any $g \in G$ with $\text{size}(g) >=2,$ split $g$ into two subgroups $g_1,g_2$ so that $$\text{size}(g_1) + \text{size}(g_2) = \text{size}(g),$$ with both $g_1,g_2$ of size at least one, and 
$C_{g}=C_{g_1}=C_{g_2}.$ Let $$G^{(1)} = \left( G \setminus \{g\} \right) \cup \{g_1,g_2\}.$$ 

Inductively, we define $G^{(0)} = G,$ and $G^{(n+1)}$ is simply a refinement of $G^{(n)}.$

Sections \S\ref{secIPA} and \S\ref{secSubgroupAlg} comprise the solution to the subgroup problem. It refines the groups $G$ until they are small enough so that they can fit into all the courses without exceeding capacity. It does not give the optimal\footnote{We wrote an integer program to find the least number of groups needed to solve the subgroup problem, but the IP appears to be computationally unsolvable. Another idea is a quadratically constrained integer program.} (least) amount of groups. 

For our instance, we have around 650 students who are split into 33 groups, of unequal size. The groups range in size from 1 to 60 students. Capacities of courses generally range from 8 to 30. After running the Subgroup Algorithm we end up with roughly 40 groups, a great improvement over potentially hundreds of tiny groups (which would make the integer program computationally unsolvable). 

The algorithm is a greedy algorithm that calls a simple bin packing integer program called IPA (below). It takes about 50 iterations to solve the subgroup problem completely, and IPA can be solved almost instantly using Gurobi, CBC, or SCIP.  

For our instance, IPA has 1289 rows, 2936 columns, and 7472 non-zeros. However, an efficient pre-solver reduces the problem to 24 rows, 136 columns, 264 nonzeros.

\subsection{Integer Program A} \label{secIPA}
Consider the integer program (IPA) below. 
 
Its inputs include the following sets: $S$ is the set of sections; $G^{(n)}$  a refinement of $G$ for some integer $n \geq 0;$ $W^{(n)}= \{ (g,s) \in G^{(n)} \times S~|~ s \in \pi^{-1}(C_g) \}.$

We need the following variables: the variables  $x_{gs},$ binary, where $(g,s) \in W^{(n)},$  equal to 1 iff group $g$ is enrolled in section $s;$ $t_s,$ non-negative integer variables, where  $s \in S,$  represents over-enrollment beyond the capacity of $s.$ 

\subsubsection{IPA}

\textbf{Objective}
\begin{align}
\min \quad
  z=  &  \sum_{s\in S} t_s
\end{align}

\textbf{Constraints}

\begin{align}
	\sum_{s \in S~:~ \pi(s)=c} x_{gs}  &= 1  && \forall g \in G^{(n)},~ \forall c \in C_g \\
	\sum_{g \in G^{(n)}~:~ (g,s) \in W^{(n)}} \text{size}(g) x_{gs} & \leq  \text{cap}(s) + t_s && \forall s  \in S. 
	\end{align}
	\begin{align}
	\forall~ g \in G^{(n)}, s_0,s_1 \in S \times S: ~ (g,s_0),(g,s_1) \in W^{(n)},  \\ \text{lab}(s_0)=\text{False},~ \text{lab}(s_1)=\text{True},~ \text{labtie}(s_0) \neq \emptyset, \text{labtie}(s_0) = \text{labtie}(s_1),  \\ x_{g_{s_{0}}} & \geq x_{g_{s_{1}}}
\end{align}

\subsection{The Subgroup Algorithm} \label{secSubgroupAlg}

Before this algorithm is run, we assume that enough sections $s \in S$ are offered to cover the demand of the groups $g \in G$ who request courses.

This algorithm calls the integer program IPA (\S\ref{secIPA})
 
\begin{algorithm}[H]
 \KwData{sets $G$ and $S$\; 
  variables\; 
  array $t$ = $(t_s)$ with $s \in S$\;
  $x$ = $x_{gs}$ with $g \in G, s \in S$\;
  $z \in \NN$ }
 \KwResult{a refinement $G^{(n)}$ of $G,$ where $n$ is some integer so that no section $s \in S$ is over capacity when all groups of $G^{(n)}$ are assigned to sections.  }
 $G^{(0)} := G$\;
 \tcc{Start with the original groups, which are too big to fit into the sections without exceeding capacity}
 $n:=0$\;
 \tcc{Next we call Integer Program A and store the optimal values}
 $z$ := optimal objective of IPA($G^{(0)}$)\;
 $t$ := optimal $t_s$-values of IPA($G^{(0)}$)\;
 $x$ := optimal $x_{gs}$-values of IPA($G^{(0)}$)\;
 
 \tcc{When $z > 0$ at least one section is over capacity}
 \While{$z > 0$}{
 	\tcc{Find the section that is most overbooked}
	 Find $r$ so that $t_r := \max_{s \in S}(t_s),$ \;
	\tcc{Find all groups that are enrolled in section $r$} 
	 Find all $g_1,g_2,\dots,g_k \in G^{(n)}$ with $x_{g_i r} \neq 0$\\
	 \tcc{Find the biggest group in over capacity section $r$}
	 choose $g = g_i$ with maximal $\text{size}(g_i)$ 
	 split $g$ into two groups, $g',g''$ where $\text{size}(g') = \text{size}(g)-t_r$ and $\text{size}(g'') =t_r$\;
	 \tcc{By optimality of $t_r,$ it follows that $\text{size}(g)-t_r > 0$}
	 set $G^{(n+1)} = (G^{(n)} \setminus \{g\}) \cup \{g',g''\} $ \;
	\tcc{Run IPA with the refined group  $G^{(n+1)}$}
	$z$ = optimal objective of IPA($G^{(n+1)}$)\;
 	$t$ = optimal $t_s$-values of IPA($G^{(n+1)}$)\;
 	$x$ = optimal $x_{gs}$-values of IPA($G^{(n+1)}$)\;
	$n = n+1$\;
}

 \caption{Subgroup Algorithm}
\end{algorithm}

\section{Formulation of the Timetabling Integer Program}

\subsection{Sets}
$Y = \{s \in S, d \in D, t \in T, r \in R:\text{roomtype}(s) = \text{type}(r), \text{cap}(s) \leq \text{cap}(r) \}.$  
This reduces the complexity of the search space since only appropriate rooms are considered. 

$W = \{ (g,s) \in G \times S: s \in \pi^{-1}(C_g) \}.$ Only certain sections are an option for each group. 

$A = \{ (a,b,c) \in T^3 : a < b < c \} \setminus \{ (1,2,3), (2,3,4), (5,6,7) \}.$ These times are not allowed for three period contiguous labs.

$B = \{ (a,b) \in T^2 : a < b \} \setminus \{(a,b) \in T^2 : a \neq 4, b = a+1 \}.$ These times are not allowed for two period labs (recall that lunch falls between periods 4 and 5). 

$H$ is the set of professors who are considered full-time, $$H = \{ p \in P : \sum_{\substack{s \in S:  \text{prof}(s)=p}} \text{periods}(s) \geq 9 \}.$$ 
\subsection{Variables}
All variables are binary unless specified otherwise.

\subsubsection{Major variables}
\begin{enumerate}
\item For $\{ (s,d,t,r) \in Y\},$  define $z_{sdtr} = 1$ if section $s$ is scheduled for day $d,$ time $t,$ and room $r,$ otherwise zero.

\item For $\{(p,d,t) \in P \times D \times T\},$ define $w_{pdt} = 1$ if professor $p$ is scheduled as the primary professor of some section that meets on day $d$ and time $t,$ otherwise zero.

\item For $\{(g,s) \in W\},$ define $x_{gs} = 1$ if group $g$ is scheduled to attend  section $s,$ otherwise zero.

\item For $\{(g,d,t,s) \in G \times D \times T \times S: (g,s) \in W \}$ define $u_{gdts} = 1$ if group $g$ is scheduled to attend  section $s$ on day $d$ at time $t,$ otherwise zero.

\end{enumerate}

\subsubsection{Auxiliary variables}
All variables are binary, 0-1, unless specified otherwise. Note that the 0-1 variables all give implicit constraints in the formulation of the IP.  The auxiliary variables' meaning will be clear from their usage in the constraint section. 

\begin{enumerate}
\item For $\{ (s,d) \in S \times D : \text{lab}(s)= \text{True} \},$  $y^1_{sd}$

\item For $\{ (s,d,r) \in S \times D \times R : \text{lab}(s)= \text{True}, ~ \text{roomtype}(s) = \text{type}(r), \text{cap}(s) \leq \text{cap}(r)\},$  $y^2_{sdt}$

\item For $\{ (p,d) \in P \times D \},$  $y^3_{pd}$

\item For $\{ (p,d) \in P \times D \},$  $t^4_{pd}$

\item For $\{ p \in P \},$  $t^\oldtext{tue}_{p}$

\item For $\{ p \in P \},$  $t^5_{p}$

\item For $\{ (s,d) \in S \times D : \text{lab}(s)= \text{False}, ~ 2 \leq  \text{periods}(s) \leq 3 \},$ $y^{\oldtext{gp1}}_{sd}$

\item For $\{ (s,d) \in S \times \{M,T,W,R \} : \text{lab}(s)= \text{False}, ~2 =  \text{periods}(s)  \},$ $t^{\oldtext{gp2}}_{sd}$

\item For $\{ (s,d) \in S \times \{M,T,W \} : \text{lab}(s)= \text{False}, ~ 3 =  \text{periods}(s)  \},$ $t^{\oldtext{gp3}}_{sd}$

\item For $\{ (p,d,t) \in P \times D \times T: \text{avail}(p)_{dt} \leq 0 \},$ $t^\oldtext{0}_{pdt}$

\item For $\{ s \in S\},$ $t_s \geq 0,$ integer

\end{enumerate}

\subsection{Formulation of the Hard Constraints}

\subsubsection{} 
Each section $s$ must be scheduled for exactly $\text{periods}(s)$ time slots per week.

\begin{align}
	\forall~ s \in S, \quad  \sum_{\substack{
	(d,t,r) \in D\times T \times R: \\ (s,d,t,r) \in Y }} z_{sdtr}  &= \text{periods}(s)  
	\end{align}

\subsubsection{}
If two sections $s_1,s_2$ have $\text{link}(s_1)=\text{link}(s_2),$ then they must be taught at identical days and times.

\begin{align}
	\forall~ s_1,s_2 \in S, t \in T, d \in D: \text{link}(s_1) = \text{link}(s_2)  \neq 0,  \\ \sum_{\substack{
	r \in R: \\ (s_1,d,t,r) \in Y }} z_{s_1dtr}  &=   \sum_{\substack{
	r \in R: \\ (s_2,d,t,r) \in Y }} z_{s_2dtr}
	\end{align}

\subsubsection{}
For each $d \in D, t \in T, r \in R$ at most one section $s$ can be assigned (one class at a time in each room).

\begin{align}
	\forall~ d \in D, t \in T, r \in R, \\ \sum_{\substack{
	s \in S: \\ (s,d,t,r) \in Y }} z_{sdtr}  &\leq 1  
	\end{align}

\subsubsection{}
Each section that has mandated times must be scheduled accordingly. 

Suppose section $s$ must meet on day $d_0,$ that is $(d_0,\emptyset) \in \text{mandate}(s),$

\begin{align}
	\forall~ s \in S, d_0 \in D, i \in \{1\dots 6\}:~ \text{mandate}_i(s) = (d_0,\emptyset),  \\ \sum_{\substack{
	t \in T, r \in R : \\ (s,d_0,t,r) \in Y }} z_{sd_0tr}  & \geq 1  
	\end{align}
The ``$\geq$'' is necessary because the section could be a multi-period lab.	

Next suppose  section $s$ must meet on day $d_0$ and time $t_0,$ that is $(d_0,t_0) \in \text{mandate}(s),$

\begin{align}
	\forall~ s \in S, d_0 \in D, t_0 \in T, i \in \{1\dots 6\}:~ \text{mandate}_i(s) = (d_0,t_0),  \\  \sum_{\substack{
	r \in R : \\ (s,d_0,t_0,r) \in Y }} z_{sd_0t_0r}  & = 1  
	\end{align}

\subsubsection{}
Each section that is not a lab can meet at most one time per day.

\begin{align}
	\forall~ s \in S, d \in D: \text{lab}(s)=\text{False}, \\ \sum_{\substack{
	t \in T, r \in R : \\ (s,d,t,r) \in Y }} z_{sdtr}  & \leq 1  
	\end{align}

\subsubsection{}
Each section that is a lab must have all its meetings on the same day, in the same room, contiguously scheduled, and without lunch falling within the allotted times. 

Either all meetings of a lab are on a day, or none are:
\begin{align}
	\forall~ s \in S, d \in D: \text{lab}(s)=\text{True}, \\  \sum_{\substack{
	t \in T, r \in R : \\ (s,d,t,r) \in Y }} z_{sdtr}  & \leq \text{periods}(s) y^1_{sd} \\
	\text{periods}(s) & \leq  \text{periods}(s)(1-y^1_{sd}) + \sum_{\substack{
	t \in T, r \in R : \\ (s,d,t,r) \in Y }} z_{sdtr} 
\end{align}

Each lab is assigned the same room over multiple periods
\begin{align}
	\forall~ s \in S, d \in D, r \in R: \text{roomtype}(s)=\text{type}(r), \\ \text{cap}(s) \leq \text{cap}(r), ~ \text{lab}(s)=\text{True}, \\   
	\sum_{ t \in T} z_{sdtr}  & \leq \text{periods}(s) y^2_{sdr} \\
	\text{periods}(s) & \leq  \text{periods}(s)(1-y^2_{sdr}) + \sum_{t \in T} z_{sdtr} 
\end{align}

Contiguous periods are required

Two period labs:
\begin{align}
	\forall~ s \in S, d \in D, r \in R, (t_1,t_2) \in B: \text{roomtype}(s)=\text{type}(r), \\ \text{cap}(s) \leq \text{cap}(r), \\ ~ \text{lab}(s)=\text{True}, \text{periods}(s) = 2  \\   
	 z_{sdt_1r} + z_{sdt_2r}   \leq 1 
\end{align}

Three period labs:
\begin{align}
	\forall~ s \in S, d \in D, r \in R, (t_1,t_2,t_3) \in A: \text{roomtype}(s)=\text{type}(r), \text{cap}(s) \leq \text{cap}(r) \\ ~ \text{lab}(s)=\text{True}, \text{periods}(s) = 3  \\   
	 z_{sdt_1r} + z_{sdt_2r} + z_{sdt_3r}   \leq 2 
\end{align}

Four period labs:
\begin{align}
	\forall~ s \in S, d \in D, r \in R : \text{roomtype}(s)=\text{type}(r),\text{cap}(s) \leq \text{cap}(r) \\  ~ \text{lab}(s)=\text{True}, \text{periods}(s) = 4  \\   
	 z_{sd5r} + z_{sd6r} + z_{sd7r}  &= 0 
\end{align}

\subsubsection{}
 For each $p \in P$ the set of sections $s$ with $\text{prof}(s)=p$ must be scheduled at disjoint times (no collisions allowed), and they must all be scheduled.  

\begin{align}
	\forall~ p \in P, d \in D, t \in T \\  \sum_{\substack{
	(s,r) \in S \times R : \\ (s,d,t,r) \in Y \\ \text{prof}(s)=p }} z_{sdtr}  & = w_{pdt} 
\end{align}
Since $w_{pdt}$ is a binary variable, collisions are avoided. 

\subsubsection{}
For each $s \in S, p \in P$ with $\text{coprof}_i(s)=p$ for some $i,$ each section $s_k$ with $\text{prof}(s_k)=p$ must be scheduled so professor $p$ can attend section $s.$ In other words, if $p$ has regular sections, they can not make it impossible for him/her to attend sections where he/she co-teaches. 

\begin{align}
	\forall~ p \in P, d \in D, t \in T, s \in S, i \in {1 \dots 6}: \text{coprof}_i(s) = p  \\  w_{pdt} & \leq 1 - \sum_{\substack{
	r \in R : \\ (s,d,t,r) \in Y}} z_{sdtr}
\end{align}

\subsubsection{}
For each group $g\in G,$ and each $c \in C_g,$ group $g$ must be scheduled for exactly one section $s \in \pi^{-1}(c).$ Note that we are not preassigning groups to sections. 
\begin{align}
	\forall~ g \in G, c \in C_g, \\  \sum_{\substack{
	s \in S : \\  s \in \pi^{-1}(c) }} x_{gs} &= 1 
\end{align}

\subsubsection{}
For each $g \in G,$ the set of sections that $g$ is enrolled in, must be scheduled at disjoint times (no collisions).

Each section $s$ that $g$ is enrolled in should be in scheduled in the timetable of $g$ for the correct amount of periods that $s$ meets
\begin{align}
	\forall~ (g,s) \in W, \quad  \sum_{\substack{(d,t) \in D\times T  }} u_{gdts} &= \text{periods}(s) x_{gs} 
\end{align}

If group $g$ is enrolled in section $s$ then reserve the periods that $s$ meets for the timetable of $g$
\begin{align}
	\forall~ (g,d,s,t) \in G \times D \times S \times T: (g,s) \in W,  \quad
	u_{gdts} + (1-x_{gs}) &\geq \sum_{\substack{r \in R: \\ (s,d,t,r) \in Y }} z_{sdtr}
\end{align}

Each group should be enrolled the correct total number of periods that the group requires
\begin{align}
\forall~ g \in G,  \quad  \sum_{\substack{(d,t,s) \in D\times T \times S: \\ (g,s) \in W }} u_{gdts} &= \sum_{\substack{c \in C_g}}\text{periods}(c) 
\end{align}

Each group should only have one section at a time\footnote{The above constraint forced us to define the variable $u_{gdts}$ with four subscripts. We spent a lot of time looking for a formulation with only three subscripts, to reduce the complexity of the problem. We did succeed, however, though we had much less variables, we needed many  extra constraints, which made the IP computationally infeasible. The necessary constraint of tying the section schedules to the group schedules seems to increase the complexity of this IP.}
\begin{align}
	\forall~ (g,d,t) \in G \times D \times T, \quad
	 \sum_{\substack{s \in S: \\ (g,s) \in W }} u_{gdts} &\leq 1  
\end{align}

\subsubsection{}
If $s_0 \in S$ is a non-lab, and $s_1 \in S$ is a lab, and $\text{labtie}(s_0) = \text{labtie}(s_1) \neq 0,$ then all groups enrolled in $s_1$ must also be enrolled in $s_0.$ This constraints keeps labs and lectures together, with the same professor, if requested. 

\begin{align}
	\forall~ g \in G, s_0,s_1 \in S \times S: ~ (g,s_0),(g,s_1) \in W, \\ \text{lab}(s_0)=\text{False},~ \text{lab}(s_1)=\text{True},~ \text{labtie}(s_0) \neq \emptyset, \text{labtie}(s_0) = \text{labtie}(s_1),  \quad x_{g_{s_{0}}} & \geq x_{g_{s_{1}}}
\end{align}

\subsubsection{}
For each section $s,$ the number of students enrolled should not exceed $\text{cap}(s).$ We sometimes make this a soft constraint, but impose a big penalty for over-capacity. We make sure that the rooms requested for $s$ have enough capacity, so we do not explicitly check room capacity. 

If this is a hard constraint then
\begin{align}
	\forall~ s \in S, \quad \sum_{\substack{g \in G: \\ (g,s) \in W }} \text{size}(g) x_{gs} \leq \text{cap}(s)  
\end{align}

If it is a soft constraint, 
\begin{align}
	\forall~ s \in S, \quad \sum_{\substack{g \in G: \\ (g,s) \in W }} \text{size}(g) x_{gs} \leq \text{cap}(s) + t_s 
\end{align}
Here $t_s$ is a non-negative integer variable. The objective would minimize $\sum_{s \in S} c^t_s t_s, $ where $c^t_s$ is an appropriate weight. For small labs, with zero room for excess capacity, $c^t_s := M,$ for some large big M, and for regular classes, a more modest weight would suffice. 

\subsubsection{}
If a group $g$ is scheduled for two sections $s_0, s_1$ with $\text{labtie}(s_0) = \text{labtie}(s_1),$ then for each day, at most four hours should be scheduled, combined, for both $s_0$ and $s_1.$ In other words, we don't want a four hour lab scheduled on the same day as the lecture it is tied to.
\begin{align}
	\forall~  (g,s_0), (g,s_1) \in W, d \in D: ~ s_0 \neq s_1,  \\ \text{labtie}(s_0) \neq \emptyset, ~ \text{labtie}(s_0) = \text{labtie}(s_1),  \quad \sum_{t \in T} (u_{gdts_0} + u_{gdts_1}) &\leq 4
\end{align}

\subsubsection{}
If a group $g$ is scheduled for two sections $s_1, s_2$ with $\text{labtie}(s_0) = \text{labtie}(s_1),$ then there should be at least a one period break (a lunch break is acceptable) between the two sections. In other words, a lab should not start right before or after lecture if they are tied together. 
\begin{align}
	\forall~ (g,s_0), (g,s_1) \in W,  d \in D, t_0,t_1 \in T: ~ t_0 \neq 4, t_1=t_0+1, ~ s_0 \neq s_1,   \\ \text{labtie}(s_0) \neq \emptyset,  \text{labtie}(s_0) = \text{labtie}(s_1),  \quad  u_{gdt_0s_0} + u_{gdt_1s_1} &\leq 1
\end{align}

\subsection{Formulation of the Soft Constraints}

\subsubsection{}
Each two or three period section $s$ that is not a lab, should not be scheduled  on consecutive days.
\begin{align}
\forall  s\in S, d \in D : \text{lab}(s)= \text{False}, ~ 2 \leq  \text{periods}(s) \leq 3, \\
y^{\oldtext{gp1}}_{sd} & = \sum_{\substack{(t,r) \in T \times R: \\ (s,d,t,r) \in Y }} z_{sdtr}
\end{align}

For $d \in \{M,T,W,R\}$ define $\text{next}(d)$ to be the day after day $d.$

When $\text{periods}(s)=2,$ 
\begin{align}
\forall  s\in S, d_0 = \{M,T,W,R\}, d_1 \in D: d_1 =\text{next}(d_0),~  \text{lab}(s)= \text{False},~  \text{periods}(s) = 2, \\ y^{\oldtext{gp1}}_{sd_0} + y^{\oldtext{gp1}}_{sd_1}  \leq & ~1 + t^{\oldtext{gp2}}_{sd_0}
\end{align}

When $\text{periods}(s)=3,$ 
\begin{align}
\forall  s\in S, d_0 = \{M,T,W\}, d_1,d_2 \in D: d_1 =\text{next}(d_0),~d_2 = \text{next}(d_1), ~  \\ \text{lab}(s)= \text{False},~  \text{periods}(s) = 3, \\ y^{\oldtext{gp1}}_{sd_0} + y^{\oldtext{gp1}}_{sd_1} + y^{\oldtext{gp1}}_{sd_2}  \leq & ~2 + t^{\oldtext{gp3}}_{sd_0}
\end{align}

The sum of the all the variables $t^{\oldtext{gp3}}_{sd}$ and $t^{\oldtext{gp2}}_{sd}$ are minimized in the objective with a small weight. When they are zero, the constraint is fully satisfied.  

For our real data, a few sections violate these soft constraints.

\subsubsection{}
Each professor must have at least one day without teaching.
\begin{align}
	\forall~ p \in P, d \in D,  \quad 
	 \sum_{t \in T} w_{pdt}  &\leq  7y^3_{pd} \\ 
	 \sum_{t \in T} w_{pdt}  &\geq  y^3_{pd} 
\end{align}
\begin{align}	 
	 \forall~ p \in P, \quad 
	 \sum_{d \in D} y^3_{pd}  & \leq  4 + t^5_p
\end{align}

The variables $t^5_p$ are minimized in the objective.

\subsubsection{}
On any given day, a professor should not teach both first period and seventh period.
\begin{align}	 
	 \forall~ p \in P, d \in D,~ t_0,t_1 \in T:~t_0=1,~t_1=7,  \quad  w_{pdt_0}+w_{pdt_1} \leq & 1 + t^4_{pd}
\end{align}
In the objective the variables $t^4_{pd}$ are minimized.

\subsubsection{}
All full-time professors (those teaching 9 or more hours per week) should be assigned teaching duties on Tuesdays (for weekly department meetings).
\begin{align}	 
	 \forall~ p \in H,~ d \in D:~d = T~,   \quad  y^3_{pd}+ t^\oldtext{tue}_{p} \geq ~&~ 1
\end{align}

In the objective, we minimize, with a small weights $d^\oldtext{tue},$  the sum $$d^\oldtext{tue} \sum_{p \in H} t^\oldtext{tue}_{p}.$$

\subsubsection{}
For each $p \in P,$ if $\text{avail}(p)_{dt} \leq 0,$ then do not schedule any section with $\text{prof}(s) = p,$ or $\text{coprof}_i(s) = p,$ at time and day $(d,t).$

For the main professor of each section
\begin{align}
 p \in P,~ d \in D, ~ t \in T: \text{avail}(p)_{dt} \leq 0, \quad w_{pdt} & \leq t^\oldtext{0}_{pdt}
\end{align}

For co-professors\footnote{Co-professors are allowed to be co-scheduled for more than one section at a time.}

\begin{align}
	\forall~ p \in P, d \in D, t \in T, s \in S, i \in {1 \dots 6}: \text{coprof}_i(s) = p  \\  \sum_{\substack{
	r \in R : \\ (s,d,t,r) \in Y}} z_{sdtr} & \leq t^\oldtext{0}_{pdt}
\end{align}

Let $c_0$ be a large weight, and let $$c_p = c_0 10^{-\text{avail}(p)_{dt}},$$ then the objective minimizes $$\sum_{\substack{(p,d,t) \in P \times D \times T: \\  \text{avail}(p)_{dt} \leq 0  }} c_p t^\oldtext{0}_{pdt}. $$

This gives an order of magnitude weight for higher priority professor-time preferences. 

\subsection{Objective}
Minimize $z$ where 
\begin{multline}
z = \sum_{\substack{(p,d,t) \in P \times D \times T: \\  \text{avail}(p)_{dt} \leq 0  }} c_p t^\oldtext{0}_{pdt} + d_4 \sum_{\substack{(p,d) \in P \times D}}t^4_{pd} + d_{\oldtext{tue}} \sum_{\substack{p \in H}}t^\oldtext{tue}_{p}  \\ +  d_{\oldtext{gp2}} \sum_{\substack{(s,d) \in S\times D: \\ d \neq F \\\text{lab}(s)= \text{False}\\ \text{periods}(s) = 2 }}t^{\oldtext{gp2}}_{sd}  + d_{\oldtext{gp3}} \sum_{\substack{(s,d) \in S\times D: \\ d \neq F \\ d \neq R \\ \text{lab}(s)= \text{False}\\ \text{periods}(s) = 3 }}t^{\oldtext{gp3}}_{sd} + d_5 \sum_{p \in P} t^5_p.
\end{multline}

Where the constants above are all weights that one chooses to weight which soft constraints should be violated above others.

\section{Computational Details}
The project started with a small, artificial model of the real timetabling problem. It was modeled with the GNU MathProg language \cite{glpk}  which is integrated with the GLPK MIP solver. When the registrar provided  one fourth of the real data, it became apparent that GLPK would not be able to solve the real problem. Since this is an academic project, we tried to solve the problem with other open source solvers. We contacted Yuji Shinano of the SCIP project \cite{scip}, but he informed us that our problem was ``quite huge'' and that we would need a cluster of 100 cores and a large amount of time to solve using ParaSCIP or FiberSCIP (parallel solvers, though not the same algorithm as SCIP). We compiled CBC \cite{cbc} to enable multithread processing, and the result was the same: no open source solver could even find a feasible solution to our IP, even with 24 hours of computational time.

We next tried commercial solvers. IBM's CPLEX \cite{cplex} was able to find feasible solutions to small versions of our real problem, but didn't succeed in finding optimal solutions, even after 24 hours and running on multiple cores. 

We next tried Gurobi 6.5 \cite{gurobi}, and the small version of the problem (roughly 1/3 of all students) was solvable in a few hours. The full version of the problem was almost solved by Gurobi 6.5 in about 5--8 hours. Often the solver was very close to the optimal solution (within a few soft constraints) and we accept the \emph{very good near optimal solution}. When Gurobi 7.0 was released, solving time was cut substantially (by more than 30 percent). 

All was going well, we were consistently solving the full problem to near optimality in 4 hours when we discovered that we left out the constraint that forces each classroom to be used by at most one lecture at a time. Coding this constraint brought the solving time to 20-40 hours. 

The next step was to tune Gurobi. We modified our small artificial problem so that it was challenging for Gurobi to solve, yet took about 1 hour. Using the automated tuning utility, we found that following parameters reduced the solving time:

\textbf{\mbox{Heuristics=0,} \mbox{FlowCoverCuts=1.}} Other, good, though less ideal parameters were \textbf{\mbox{NormAdjust=1,}} \textbf{\mbox{PreDual=0,}} were found by tuning an easier artificial problem. Combining the two sets of parameters offered no improvement.  

\subsubsection{Time to find solution} 
One of the drawbacks to using huge integer programs to solve timetabling problems is the variability of the solution time. The solver uses a random seed value to make arbitrary decisions, and this seed can have a big impact on solution times. With a large number of computer cores, one can try out many different seeds simultaneously. 

Another aspect that can greatly affect solution time are the weights on the objective.

The computer we used was a high performance laptop with an i7 Intel CPU with 4 cores, capable of executing 8 threads per cycle. The memory of the machine was 16GB and when solving with 7 threads, about 3 GB were actually used when the search tree became large. For future work we are hoping for a machine with many more cores, and an error correcting code memory CPU, such as the Xeon class of processors. 

We were surprised that such a large IP could be solved on a portable computer. 

\section{Future Work}
There are many directions this work can be extended. USMMA is in the process of transitioning to the TIP solution. We expect many new requests for enhancements over the coming years. The following are some issues on the horizon:

\subsection{Decreasing solving time} In our model, the TIP assigned actual rooms to sections. We could separate out the room assignment by simply requiring the main TIP not to assign more sections at a certain time, that need a certain room type, than are available in that class of room types. 

One could also experiment with pre-assigning students to sections. Then one only needs to assert that the sections do not collide and that the professors have no collisions. This would be a suboptimal approach, so we did not pursue it.

Our subgroup generating algorithm does not find the minimal number of groups needed to pack all groups into classes. We tried solving this subgroup problem with a 0-1 MIP and the computation time exceeded 12 hours. Another approach is a quadratically constrained integer program. We do not know if that would be faster.

\subsection{Recalculate timetables} Suppose the registrar wants to change a few schedules after the term has started. In theory we can solve the problem by modifying our objective and adding penalties for deviations of our binary variables. However, we need to study that type of changes that are requested (in real life) and see how it changes our sets and data, and understand how the TIP changes. A possible solution is to schedule dummy sections and dummy groups which may give flexibility to change our initial TIP without changing too much of the key structures. 

\subsection{Validity of data} While it is straightforward to correct simple mistakes in the real data, sometimes one can not catch a mistake until the solver returns the dreaded \emph{infeasible problem.} Some problems that we discovered were when too many sections requested rooms for a class of room types that was too small; professors were mandating their courses be scheduled at conflicting times; and 3 credit courses were requesting 4 time slots. The infeasibility problem is a serious problem and we had to compute irreducible infeasible sets (IIS) in order to track down subtle errors in the data. We are currently writing traditional computer programs to search for common errors that cause the TIP to be infeasible.

\bibliography{timeNov25}
\bibliographystyle{acm}

\vspace{5mm}

\noindent
Joshua S. Friedman \\
Professor of Mathematics \\
\textsc{United States Merchant Marine Academy} \\
300 Steamboat Road \\
Kings Point, NY 11024 \\
U.S.A. \\
e-mail: friedmanJ@usmma.edu, crowneagle@gmail.com

\end{document}